\documentclass[10pt,twocolumn,letterpaper]{article}

\usepackage{cvpr}              

\usepackage{graphicx}
\usepackage{amsmath}
\usepackage{amssymb}
\usepackage{booktabs}

\usepackage[T1]{fontenc}
\usepackage[latin9]{inputenc}
\usepackage{babel}
\usepackage{array}
\usepackage{float}
\usepackage{multirow}
\usepackage{amsbsy}
\usepackage{amstext}
\usepackage{wasysym}
\usepackage[noend]{algpseudocode}

\usepackage{algorithmicx,algorithm}

\newcommand{\lyxdot}{.}

\usepackage[pagebackref,breaklinks,colorlinks]{hyperref}

\usepackage[capitalize]{cleveref}
\crefname{section}{Sec.}{Secs.}
\Crefname{section}{Section}{Sections}
\Crefname{table}{Table}{Tables}
\crefname{table}{Tab.}{Tabs.}

\begin{document}

\title{Dynamic Background Reconstruction via MAE for Infrared Small Target Detection}

\author{Jingchao Peng, Haitao Zhao, Kaijie Zhao, Zhongze Wang, and Lujian Yao\\
Automation Department, School of Information Science and Engineering,\\ 
East China University of Science and Technology\\
}
\maketitle

\begin{abstract}
  Infrared small target detection (ISTD) under complex backgrounds is a difficult problem, for the differences between targets and backgrounds are not easy to distinguish. 
  Background reconstruction is one of the methods to deal with this problem.
  This paper proposes an ISTD method based on background reconstruction called Dynamic Background Reconstruction (DBR). 
  DBR consists of three modules: a dynamic shift window module (DSW), a background reconstruction module (BR), and a detection head (DH). 
  BR takes advantage of Vision Transformers in reconstructing missing patches and adopts a grid masking strategy with a masking ratio of 50\% to reconstruct clean backgrounds without targets.
  To avoid dividing one target into two neighboring patches, resulting in reconstructing failure, DSW is performed before input embedding. 
  DSW calculates offsets, according to which infrared images dynamically shift. 
  To reduce False Positive (FP) cases caused by regarding reconstruction errors as targets, DH utilizes a structure of densely connected Transformer to further improve the detection performance. 
  Experimental results show that DBR achieves the best F1-score on the two ISTD datasets, MFIRST (64.10\%) and SIRST (75.01\%).
\end{abstract}

\section{Introduction}

\begin{figure}[!th]
\begin{centering}
\subfloat[\label{fig:1a}The structure of the naive background reconstruction method (NBR).]{\begin{centering}
\includegraphics[width=1.0\columnwidth]{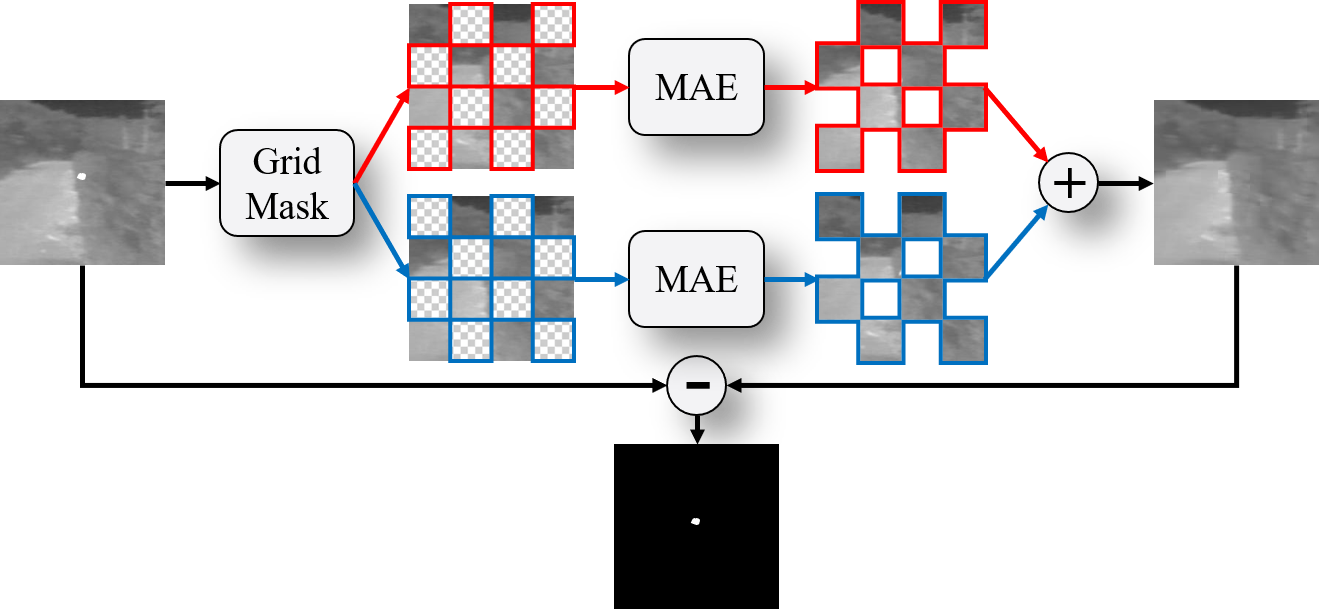}
\par\end{centering}
}
\par\end{centering}
\begin{centering}
\subfloat[\label{fig:1b}Dividing one target into two patches
will cause reconstruction failure. The probability of dividing one
target into two patches at different target sizes]{\begin{centering}
\includegraphics[width=1.0\columnwidth]{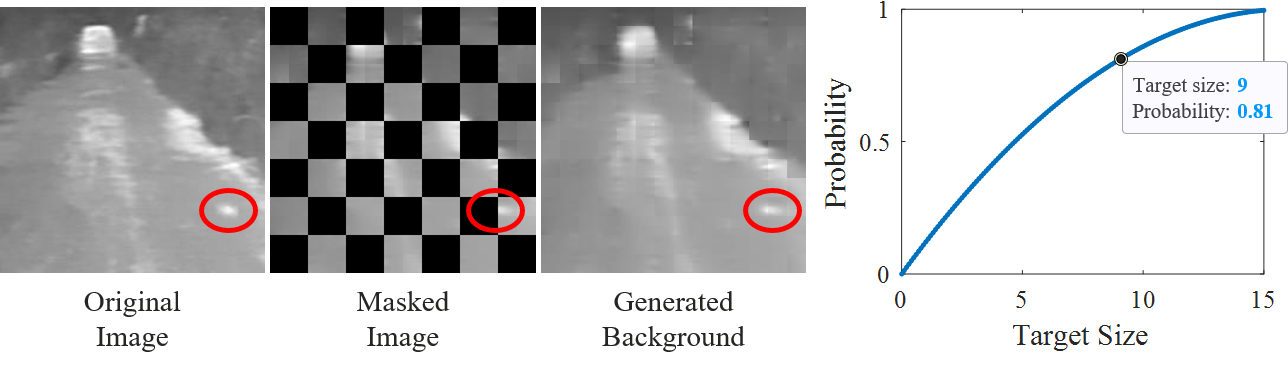}
\par\end{centering}
}
\par\end{centering}
\caption{An ISTD method based on background reconstruction and its shortage.}
\end{figure}

In recent years, infrared small-target detection (ISTD) is becoming more and more popular \cite{DNANet, IAANet, ACM, MDvsFAcGAN}.
Compared with visible light imaging, infrared imaging offers strong anti-interference, has the ability of long-range imaging, and can work around the clock \cite{survey}. 
Infrared imaging has been widely used in various fields, especially in intelligent robotics, autonomous driving, fire warning, agricultural production, leakage measurement, etc. \cite{survey, MMfusion2, MSfusion1, MSfusion3}. 
When the target is more than 10 kilometers away from the infrared detector, it usually occupies a small region (the target size is less than 9$\times$9 pixels in 256$\times$256 images \cite{smallsize}).
In addition, due to atmospheric scattering and refraction, optical defocusing, and various noises, infrared images have a low signal-to-noise (SNR) ratio and low contrast with the background \cite{noise}. 
These factors reduce the performance of ISTD methods, and thus ISTD is still a challenge in the field of target detection.

Due to extremely small sizes and low SNR ratio, targets in ISTD have inadequate detail information such as contour, shapes, textures, and so on, which is what visible light target detection methods rely on \cite{MMfusion1,MMfusion2,MMfusion3}.
Therefore, most ISTD methods pay attention to differences between targets and backgrounds.
For example, filter-based methods detect local gray differences between targets and backgrounds \cite{filter1,filter2}; human-visual-system-based (HVS) methods measure local contrast to detect targets \cite{DoG,ILCM,LCM,TLLCM,WSLCM};
deep-learning-based (DL-based) methods utilizing multiple nonlinear layers transform raw images into high-dimensional representations, in which targets and backgrounds have more significant differences than in raw images \cite{purning1,purning2,purning3,FPN1,FPN2,FPN3}.
However, given complex backgrounds, the differences between targets and backgrounds are not easy to distinguish, resulting in poor detection performance.
Consequently, detecting small targets under complex backgrounds is a difficult problem.

Clean backgrounds without targets can alleviate the detection problem under complex backgrounds \cite{BRbasedISTD}.
Since targets do not belong to backgrounds, if the regions of the targets are masked, clean backgrounds can be obtained by image inpainting \cite{inp2detect}, whose purpose is to produce visually plausible structure and texture for the missing regions of images \cite{inpainting}.
In the last few years, the success of Vision Transformers \cite{VIT} has brought new opportunities to image inpainting.
Vision Transformers embed one input image into 16$\times$16 patches.
Compared with convolutional neural networks, which process the whole feature map, Vision Transformers have natural advantages in integrating mask tokens into images. 
MAE \cite{MAE} removes random patches to reconstruct pixels under a high masking ratio (75\%) and works well. 
Therefore, the idea of masking patches with targets and reconstructing backgrounds via MAE naturally comes to us.

Inspired by MAE \cite{MAE}, we propose a naive background reconstruction method (NBR).
NBR adopts a grid masking strategy and reconstructs one image twice with a masking ratio of 50\%.
Two mask patches compose a ``mutually exclusive and collectively exhausting'' (MECE) image.
The structure of NBR can be shown in Fig. \ref{fig:1a}.
NBR assumes that masked backgrounds can be reconstructed while masked targets cannot.
The principle is that given twice reconstruction with a masking ratio of 50\%, the probability of the target being masked is 100\%.
So that NBR can reconstruct clean backgrounds without targets.
However, when a target is divided into two neighboring patches, one-half of the target will be reconstructed based on the other half of the target, resulting in the failure of background reconstruction, as shown in Fig. \ref{fig:1b}.
Furthermore, when the target size is 9$\times$9, the probability of dividing one target into two patches is up to 81\%.
It is necessary for ISTD to avoid one target being divided into two neighboring patches.

The existing methods of avoiding dividing one target into two neighboring patches include enlarging the patch window \cite{T2T-ViT,PVT-V2}, adopting deformable embedding methods \cite{PS-ViT,DAT}, and shifting windows \cite{Swin}.
However, these methods are not designed for background reconstruction and cannot be directly used in ISTD.
In specific, the methods of avoiding dividing one target into two neighboring patches, which can be used to reconstruct backgrounds, need to meet the following conditions:
\begin{itemize}
\item The same target can only be presented in one patch. Otherwise, the target will also be reconstructed.
\item The principle of MECE must be satisfied between patches. Otherwise, the background will be reconstructed incompletely.
\end{itemize}
Therefore, it is necessary to design a special method for ISTD to avoid dividing one target into two patches.

\begin{figure*}[!t]
\begin{centering}
\includegraphics[width=1\textwidth]{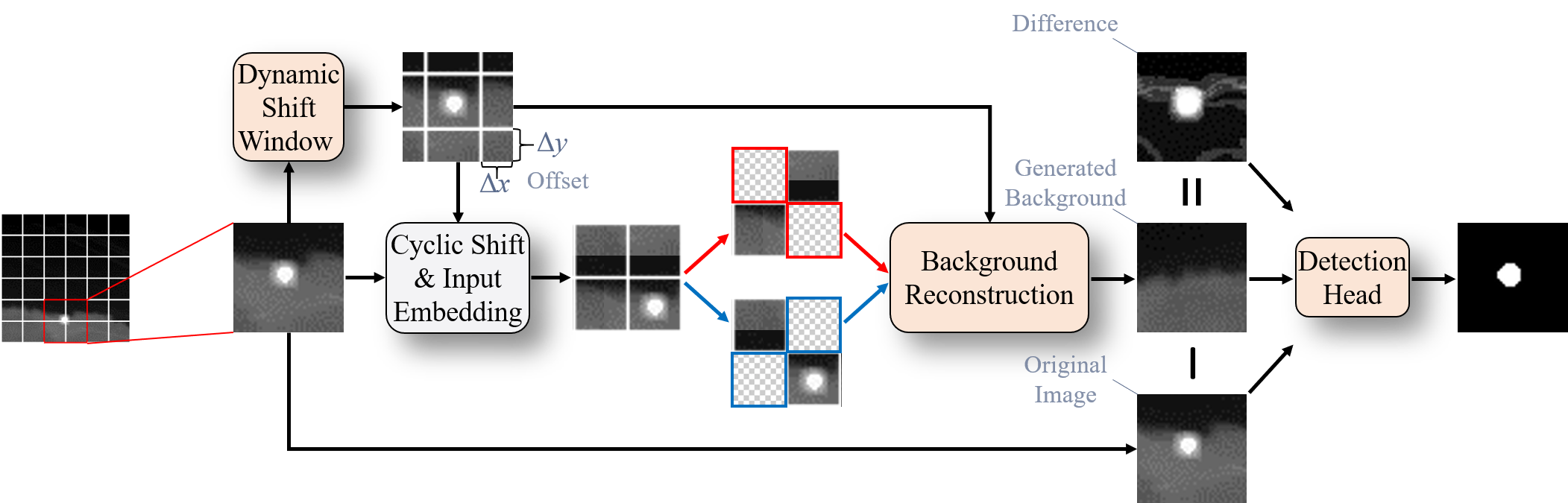}
\par\end{centering}
\caption{\label{fig:2}The architecture of the Dynamic Background Reconstruction method (DBR).}
\end{figure*}
Motivated by the above analysis, in this paper, we propose a background-reconstruction-based ISTD method called Dynamic Background Reconstruction (DBR), as shown in Fig. \ref{fig:2}.
DBR consists of a dynamic shift window module (DSW), a background reconstruction module (BR), and a detection head (DH).
First, DSW calculates an offset base on how many pixels the target can move to the center of the patch in raw infrared images.
Then, according to this offset, BR dynamically shifts windows before input embedding and uses NBR to reconstruct clean backgrounds.
Finally, infrared images with targets, clean backgrounds without targets, and their difference are concatenated and fed into DH to improve the detection performance.
Because the detector is prone to regard reconstruction errors as targets, the recall rate is greater than the precision rate.
We propose a weighted dice loss (WDLoss) to balance the precision and recall rates.

In summary, our contributions are summarized below:
\begin{enumerate}
\item DBR is a background-reconstruction-based ISTD method that can improve detection performance under complex backgrounds. 
We propose a BR based on Vision Transformer (MAE) to reconstruct clean backgrounds without targets.
\item DBR can prevent the transformer from dividing one target into two neighboring patches, which is harmful to background reconstruction.
We propose a DSW to calculate offsets, according to which BR dynamically shifts images before input embedding.
\item DBR is robust to reconstruction error.
We propose a DH and a WDLoss, which can separately reduce the influence of reconstruction error on detection performance from aspects of the network architecture and loss function.
\end{enumerate}

\section{Related Works}

In this section, we first summarize ISTD methods, including traditional and DL-based methods. 
Then we introduce existing vision transformers and how they avoid dividing one target into two patches.

\subsection{Infrared Small-Target Detectio Methods} 

ISTD methods can be largely divided into traditional methods and deep learning-based methods \cite{survey}.
Traditional methods utilize filters or the human visual system to detect targets, assuming the target is brighter than its neighbor pixels.
To strengthen the difference between targets and backgrounds, WSLCM \cite{WSLCM} adopted a weighting function and strengthened LCM \cite{LCM}; AADCDD \cite{ADDCDD} took advantage of local gray differences and employed weighting coefficients; ADMD \cite{ADMD} used directional information. 

As for DL-based methods, due to small targets prone to disappear in forward propagation, the key technology is to enhance targets in complex backgrounds.
For example, ACM \cite{ACM} and ALCNet \cite{ALCNet} utilized a bottom-up local attention module to transfer context information; 
IAANet \cite{IAANet} leveraged a transformer encoder to obtain interior relations between pixels; 
DNANet \cite{DNANet} proposed a dense nested attention network to maintain target information in deep layers; 
MDvsFA-cGAN \cite{MDvsFAcGAN} and CourtNet \cite{CourtNet} utilized two subnetworks to separately enhance targets and suppress backgrounds.
Existing ISTD methods rely on differences between targets and backgrounds to detect targets but cannot reconstruct clean backgrounds without targets.

\subsection{Vision Transformers} 

MAE \cite{MAE} developed an asymmetric encoder-decoder architecture, which masks random patches of the input image and reconstructs the missing pixels.
The reconstruction performance is excellent when the masking ratio is less than 75\%.
However, since MAE uses a fixed input embedding method, which divides the image into 16$\times$16 patches, targets are easily divided into two patches. 

To avoid dividing one target into two patches, T2T-ViT \cite{T2T-ViT} and PVT-V2 \cite{PVT-V2} utilized an overlapping patch embedding method.
The overlapping patch embedding method enlarges the patch window, and the input image is split into patches with overlapping.
Nevertheless, this method makes the target present in two patches simultaneously, so it is impossible to reconstruct a clean background based on one patch with the target.
PS-ViT \cite{PS-ViT} and DAT \cite{DAT} build deformable input embedding methods, focusing the model on complete regions.
But this method will destroy the complementation between patches, breaking the principle of MECE.
SwinTransformer \cite{Swin} adopted a shifted windowing scheme to connect cross-window patches.
However, the possibility of dividing one target into two patches is doubled with two types of shifted windows switchover.
It is adverse to background reconstruction for ISTD.

\section{Proposed Method}

In this section, we first overview the architecture of DBR. 
Then we introduce the main modules: DSW, BR, and DH.

\subsection{Overall Structure}

The architecture of DBR can be shown in Fig. \ref{fig:2}. 
Taking an infrared image as an example, if the image is directly embedded, the target will be divided into different patches.
To solve this problem, in DBR, the image is first fed into DSW to calculate an offset $(\Delta x,\Delta y)$.
The offset $(\Delta x,\Delta y)$ means that when the image horizontally shifts $\Delta x$ pixels and vertically shifts $\Delta y$ pixels, the target moves to the center of a patch instead of being divided into different patches.
In the background reconstruction phase, after input embedding, the infrared image was masked twice in a complementary way.
The masking strategy is grid masking.
Theoretically, the target can be obtained by subtracting the generated background from the original image.
However, because the generated background is not absolutely the same as the factual background, there is inevitably a reconstruction error.
To minimize the influence of the reconstruction error on detection performance, finally, the original image, the generated background, and their difference are concatenated and fed into DH.

\subsection{Dynamic Shift Window Module}

\subsubsection{Start from Single-Target}

\begin{figure}[!h]
\begin{centering}
\includegraphics[width=0.5\textwidth]{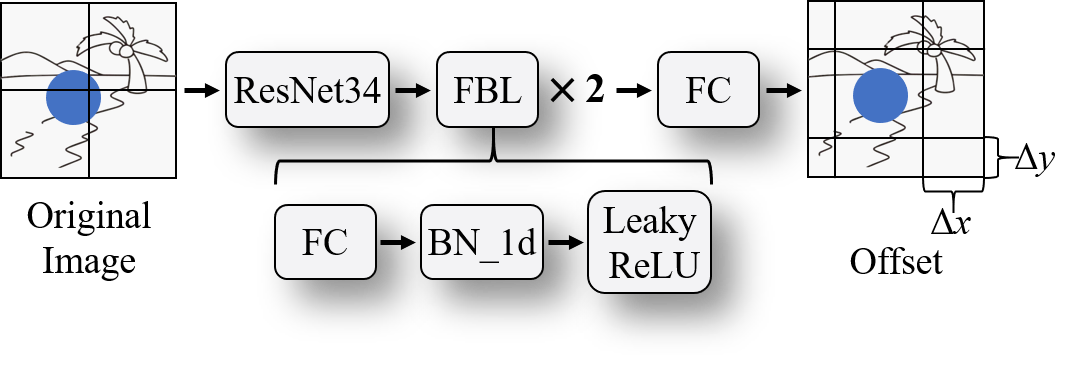}
\par\end{centering}
\caption{\label{fig:3}The structure of the dynamic shift window module (DSW).
The blue circle represents the target, and the sketch represents the background.}
\end{figure}

The structure of DSW can be shown in Fig. \ref{fig:3}.
To distinguish the target and the background, in Fig. \ref{fig:3}, Fig. \ref{fig:4}, and Fig. \ref{fig:5}, the blue circle represents the target, and the sketch represents the background.
DSW adopts ResNet34 as the backbone and three fully-connected layers as the head.
Given an infrared images $X$, the backbone $\mathcal{B}(\cdot)$, and head $\mathcal{H}(\cdot)$, the offset vector $[\Delta\boldsymbol{x},\Delta\boldsymbol{y}]$ can be obtained by:
\begin{equation}
[\Delta\boldsymbol{x},\Delta\boldsymbol{y}]=\mathcal{H}(\mathcal{B}(X)),
\end{equation}
where $\Delta\boldsymbol{x},\Delta\boldsymbol{y}\in\mathbb{R}^{16}$, indicates the probability of pixels the target should move horizontally or vertically within a patch (patch size is 16).

In the test phase, the offset can be obtained by:
\begin{equation}
\Delta x,\Delta y=(24-\textrm{argmax}([\Delta\boldsymbol{x},\Delta\boldsymbol{y}]))\%16,
\end{equation}
where $\textrm{argmax}([\Delta\boldsymbol{x},\Delta\boldsymbol{y}])$ is the target center.

In the training phase, DSW treats the offset calculation as a fitting task.
For any target, the distance from the target to its nearest patch center is $(cx,cy)$.
The distance vector $(\boldsymbol{cx},\boldsymbol{cy})$ can be obtained by rounding to the nearest integer and one-hot encoding:
\begin{equation}
\boldsymbol{cx},\boldsymbol{cy}=\textrm{onehot}(\textrm{round}(cx,cy)).
\end{equation}
DSW fits the distance vector and the offset vector with MSE loss. The specific process of generating distance vector can be shown in Alg. \ref{alg:2}.

\begin{algorithm}[!h]
\small
\caption{\label{alg:2}The distance vector generation for a single target.} 
\hspace*{0.02in} {\bf Input:}
The binary image: $X_{gt}\in\mathbb{R}^{208\times208}$.\\
\hspace*{0.02in} {\bf Output:}
The distance vector: $(\boldsymbol{cx},\boldsymbol{cy})$.
\begin{algorithmic}[1]
\State Obtain the target center $(x,y)$: \[(x,y)=\textrm{cv.findcounters}(X_{gt}),\] where cv.findcounters is the function finding white objects from a black background in OpenCV;
\State Calculate the distance from the target to its nearest patch center: \[(cx,cy)=(8-x\%16,8-y\%16),\]where ``\%'' means modulo operation;
\State Obtain the distance vector: \[(\boldsymbol{cx},\boldsymbol{cy})=\textrm{onehot}(\textrm{round}(cx,cy)).\]
\end{algorithmic}
\end{algorithm}

\begin{figure}[!t]
\begin{centering}
\includegraphics[width=0.175\textwidth]{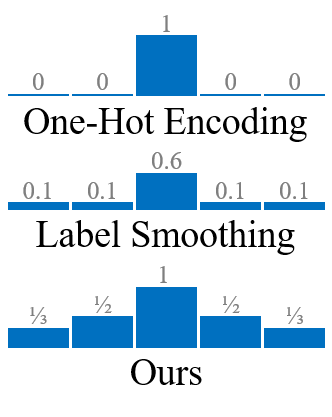}
\par\end{centering}
\caption{\label{fig:3.2}The difference between one-hot encoding, label smoothing, and ours.}
\end{figure}

Since the center vector is directional, i.e., the closer the target is to the center, the better the effect of avoiding one target into two patches is.
Therefore, we improve one-hot encoding by adding directional information. 
The difference between the proposed encoding and one-hot encoding can be shown in Fig. \ref{fig:3.2}.

\subsubsection{Extension to Multi-Targets}

For multi-targets in one image, shifting the infrared image cannot guarantee all target centers are close to their patch centers.
However, our purpose is to avoid dividing one target into different patches, if all targets are located in different but single patches, whether all target centers are close to their patch centers is not essential.
So we change to another criterion which is shifting the infrared image to make sure target centers are far from their path edges.

Compared with DSW for a single target, in the training phase, DSW also fits the distance vector and the offset vector with MSE loss. 
But the process of generating distance vector is quite different: for there are more than one target, multiple distance vectors can be generated. 
the process of generating distance vector for multi-targets averages and weights all distance vectors according to their target radius, which can be shown in Alg. \ref{alg:3}, Step $3\sim 5$.

\begin{figure}[!t]
\begin{centering}
\includegraphics[width=0.425\textwidth]{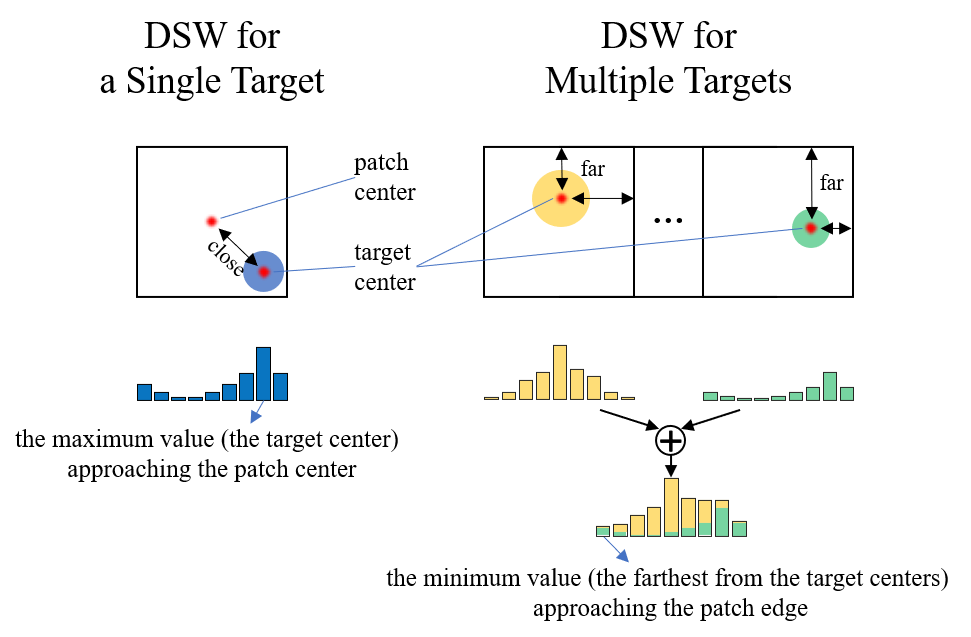}
\par\end{centering}
\caption{\label{fig:8}The difference between DSW for a single target and multiple targets.}
\end{figure}

In the test phase, the offset can be obtained by:
\begin{equation}
\Delta x,\Delta y=16-\textrm{argmin}([\Delta\boldsymbol{x},\Delta\boldsymbol{y}]),
\end{equation}
where $\textrm{argmin}([\Delta\boldsymbol{x},\Delta\boldsymbol{y}])$ is the pixel that is the farthest from the target centers.
That means for multi-targets, DSW maximizes the distance between target centers and patch edges.
The difference between the DSW of a single target and multiple targets can be intuitively shown in Fig. \ref{fig:8}.
It is worth noting that when there is only one target in an infrared image, the target center being far from the path edges is equivalent to the target center being close to the path center.
So there is no need to determine the number of targets in the test image, we directly use DSW for multiple targets (as shown in Fig. \ref{fig:8}).

\begin{algorithm}[!t]
\small
\caption{\label{alg:3}The distance vector generation for multi-targets.} 
\hspace*{0.02in} {\bf Input:}
The binary image: $X_{gt}\in\mathbb{R}^{208\times208}$.\\
\hspace*{0.02in} {\bf Output:}
The distance vector: $(\boldsymbol{cx},\boldsymbol{cy})$.
\begin{algorithmic}[1]
\State Obtain the target centers $(x_{i},y_{i})$ and their radius $r_{i}$: \[(x_{i},y_{i}), r_{i}=\textrm{cv.findcounters}(X_{gt}), i=1,2,\dots,n,\] where cv.findcounters is the function finding white objects from a black background in OpenCV;
\State Calculate the distance from the target to its nearest patch center: \[(cx_{i},cy_{i})=(8-x_{i}\%16,8-y_{i}\%16),\]where ``\%'' means modulo operation;
\For{$i$ \textnormal{in} [1,2,\dots,n]}
\State Obtain the distance vectors: \[(\boldsymbol{cx_{i}},\boldsymbol{cy_{i}})=\textrm{onehot}(\textrm{round}(cx_{i},cy_{i})).\]
\EndFor
\State Average and weight all distance vectors: \[(\boldsymbol{cx},\boldsymbol{cy})=\frac{\Sigma((\boldsymbol{cx_{i}},\boldsymbol{cy_{i}})\times r_{i})}{\Sigma(r_{i})};\]
\end{algorithmic}
\end{algorithm}

\subsection{Background Reconstruction Module}

\begin{figure*}[!t]
\begin{centering}
\includegraphics[width=1\textwidth]{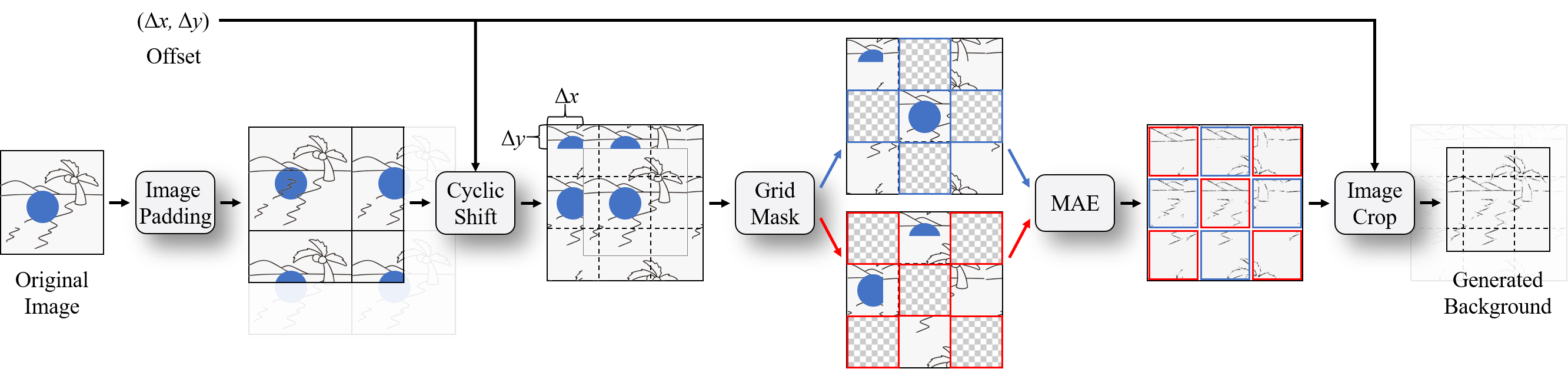}
\par\end{centering}
\caption{\label{fig:4}The structure of the background reconstruction module (BR).}
\end{figure*}

The structure of BR can be shown in Fig. \ref{fig:4}.
Since the image needs to be shifted, the down and right parts of the image will be separately shifted to the top and left, which makes the continuous image disconnected, resulting in significant differences between the two sides of the boundary.
To deal with this problem, the infrared image is first padded to ensure no boundary after shifting.
The image is then shifted according to the offset calculated by DSW, moving the target to the center of the patch.
After that, the infrared image is masked twice with the grid masking strategy.
The masking ratio is 50\%.
The two masked images are complementary.
In other words, they satisfy the principle of MECE.
After masking, MAE reconstructs the masked part.
Finally, the background of the original image size can be generated by cropping according to the offset calculated by DSW.
Given an original image $X_{ori}\in\mathbb{R}^{208\times208}$, the algorithm for reconstructing its background can be shown in Alg. \ref{alg:1}.

\begin{algorithm}[h]
\small
\caption{\label{alg:1}Background reconstruction.} 
\hspace*{0.02in} {\bf Input:}
The original image: $X_{ori}\in\mathbb{R}^{208\times208}$;\\ the offset: $(\Delta x,\Delta y)$.\\
\hspace*{0.02in} {\bf Output:}
The generated background: $X_{gen}\in\mathbb{R}^{208\times208}$.
\begin{algorithmic}[1]
\State Pad the left and top regions of $X_{ori}$ with a width of 16 pixels to the right and bottom: $X_{ori}\rightarrow X_{ori}^{\ast}\in\mathbb{R}^{224\times224}$;
\State Horizontally shift the padded image $X_{ori}^{\ast}$ by $\Delta x$ pixels, and vertically shift by $\Delta y$ pixels;
\State Embed $X_{ori}^{\ast}$ to the feature: $X_{ori}^{\ast}\rightarrow F_{ori}\in\mathbb{R}^{B\times P\times C}$;
\State Complementarily grid-mask $F_{ori}$ twice with a masking ratio of 50\%: $F_{ori}\rightarrow(F_{1},F_{2}),F_{i}\in\mathbb{R}^{B\times P\times\frac{C}{2}},F_{1}\cup F_{2}=F_{ori}$;
\For{$F_{i}$ \textnormal{in} $(F_{1},F_{2})$}
\State Reconstruct $F_{i}$ by MAE \cite{MAE}: $F_{i}\rightarrow F_{i}^{C}\in\mathbb{R}^{B\times P\times\frac{C}{2}}$;
\EndFor
\State Combine the reconstructed patch to generate the background: $F_{gen}=F_{1}^{C}\cup F_{2}^{C}$;
\State De-embed the feature to the image: $F_{gen}\rightarrow X_{gen}^{\ast}\in\mathbb{R}^{224\times224}$;
\State Crop the image to get the generated background: $X_{gen}=X_{gen}^{\ast}[\Delta x:\Delta x+208,\Delta y:\Delta y+208]$.
\end{algorithmic}
\end{algorithm}

\begin{figure}[!ht]
\begin{centering}
\includegraphics[width=1\columnwidth]{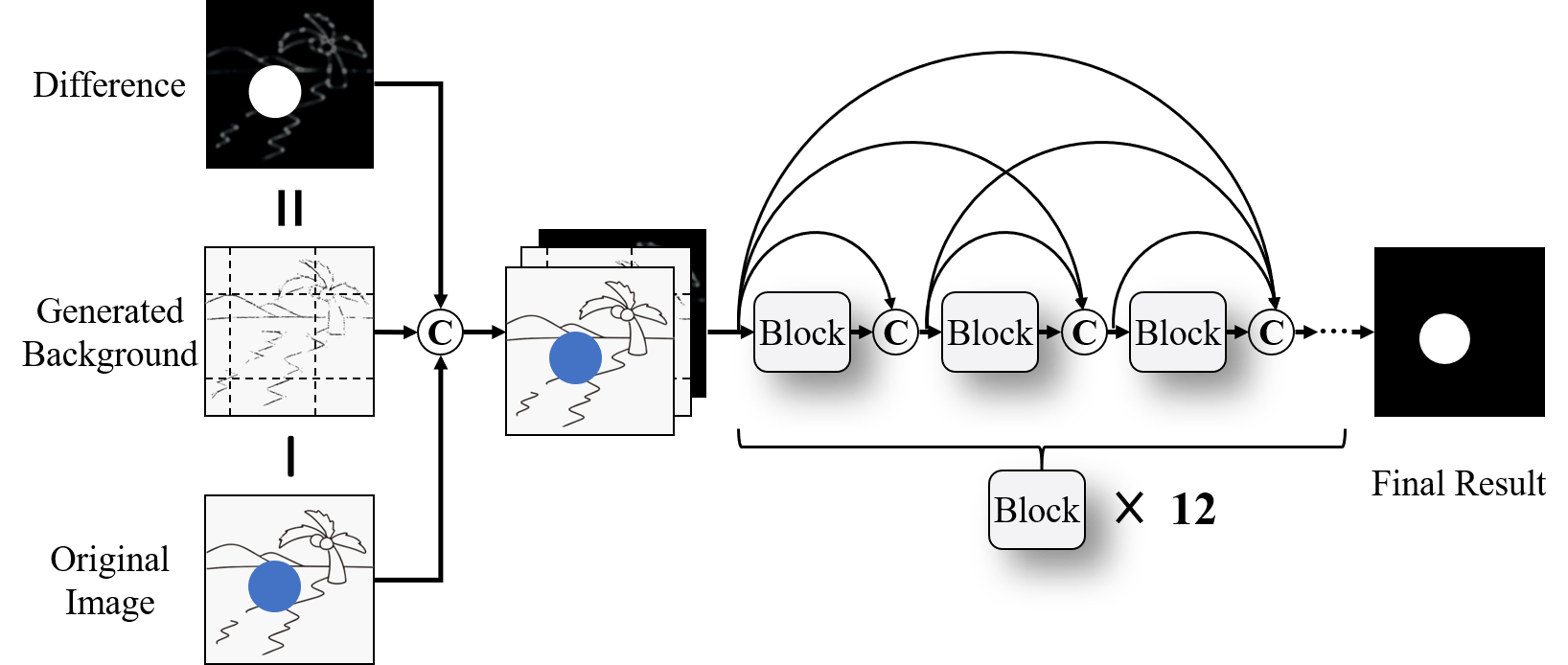}
\par\end{centering}
\caption{\label{fig:5}The structure of the detection head (DH).}
\end{figure}

\subsection{Detection Head}

The structure of DH can be shown in Fig. \ref{fig:5}.
Because the generated background is not absolutely the same as the factual background, there is inevitably a reconstruction error.
The purpose of DH is to reduce the influence of the reconstruction error on detection performance.
Inspired by CourtNet \cite{CourtNet}, DH utilizes a densely connected transformer structure consisting of 12 blocks.
Consider an original image $X_{\textrm{ori}}\in\mathbb{R}^{B\times1\times H\times W}$ and a generated background $X_{\textrm{gen}}\in\mathbb{R}^{B\times1\times H\times W}$, the batch size is $B$.
Each image has one gray channel, and its width and height are $W$ and $H$, respectively.
First, $X_{\textrm{ori}}$, $X_{\textrm{gen}}$, and their difference are concatenated and encoded to the feature $F\in\mathbb{R}^{B\times P\times C}$, where $P=p\times p$, $C=\frac{H}{p}\times\frac{W}{p}\times c$.

Then the feature $F$ goes through 12 blocks to get the outputs $Y\in\mathbb{R}^{B\times1\times H\times W}$, which are binary maps with 0 and 1 indicating whether each pixel is a target.
Suppose the input of the $i$-th block is $F_{i-1}\in\mathbb{R}^{B\times P\times C_{i-1}}$, the output is $\mathcal{F}(F_{i-1})\in\mathbb{R}^{B\times P\times32}$.
Then the input feature and output feature are concatenated:
\begin{equation}
F_{i}=\textrm{concat}(F_{i-1},\mathcal{F}(F_{i-1}))\in\mathbb{R}^{B\times P\times C_{i}},
\end{equation}
where $C_{i}=C_{i-1}+32$.
For ViT, the input feature dimension of each block is equal to the output feature dimension, and the input feature is not preserved.
The output feature dimension of our block is 32.
The input feature is concatenated with the output feature.
The advantage of concatenation is to 1) reduce the amount of computation and increase the detection speed; 2) preserve generated backgrounds, so that the detector can take full advantage of the background-reconstruction-based method.

Since the difference of $X_{\textrm{ori}}$ and $X_{\textrm{gen}}$ contains targets (TP) and reconstruction error (FP), DH is prone to have a high recall rate and a low precision rate.
To balance the precision and recall rates, inspired by Generalised Dice Loss \cite{GDLoss}, we propose a weighted dice loss (WDLoss).
Given the result of DH as $Y$ and the ground truth as $\hat{Y}$, WDLoss can be obtained by:
\begin{equation}
\mathcal{L}_{\textrm{WDL}}=-\log(\frac{2\gamma\Sigma(Y\times\hat{Y})}{\Sigma(Y)+\gamma\Sigma(\hat{Y})}),
\end{equation}
where $\gamma$ means the weighting factor.
When $\gamma>1$, the recall rate is greater than the precision rate, while $0<\gamma<1$, the precision rate is greater than the recall rate, here we set $\gamma=2\textrm{e}-3$.

\section{Experiments}
We train and test DBR on the PyTorch platform with I7-10700K CPU and RTX TITAN GPU.
We use two ISTD datasets, MFIRST \cite{MDvsFAcGAN} and SIRST \cite{ACM}, to train and evaluate DBR.
MFIRST contains 9900 training images and 100 test images, and SIRST contains 341 training images and 86 test images.
For training settings, we utilize Adam and warm-up to train DBR.
The warm-up steps are 200, the beginning learning rate is 4e-5, the max learning rate is 2.5e-4, and the number of epochs is 150.
More detail can be referred to https://github.com/PengJingchao/DBR.

As for evaluation metrics, we use the precision rate, the recall rate, and F1-score to evaluate the binary segmentation result, which is the same as \cite{MDvsFAcGAN}.
The precision rate, the recall rate, and F1-score are calculated by:
\begin{equation}
Precision=\frac{TP}{TP+FP},
\end{equation}
\begin{equation}
Recall=\frac{TP}{TP+FN},
\end{equation}
\begin{equation}
F1-Score=\frac{2\times Precision\times Recall}{Precision+Recall}.
\end{equation}
Note that a high F-1 score indicates good performance rather than only achieving high precision or recall rates \cite{MDvsFAcGAN}.

\subsection{The Precision of DSW}

\begin{figure}[!h]
\begin{centering}
\includegraphics[width=0.5\textwidth]{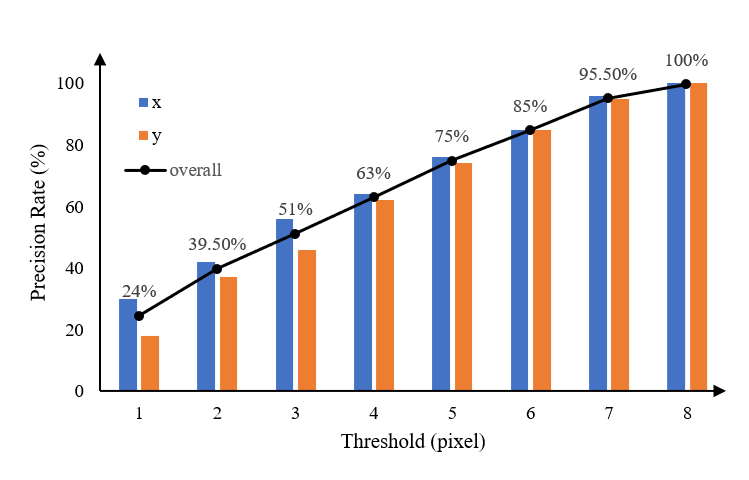}
\par\end{centering}
\caption{\label{fig:9}The precision rate of DSW under different thresholds.}
\end{figure}

The precision of DSW is crucial for background reconstruction, for if DSW cannot find the real target centers, offsets may be wrong, resulting in dividing one target into different patches.
To this end, we calculate the precision rate under the different thresholds. 
The results can be shown in Fig. \ref{fig:9}.
From the figure, we can find that 1) all target centers can be located within 8 pixels; 2) DSW can find target centers within 5 pixels with a precision rate of 75\%.
Considering the small target size ($9\times 9$) and patch size ($16\times 16$), this result is acceptable.

However, good performance does not mean that using DSW alone can satisfy the ISTD problem.
We change a head of DSW that uses an upsampling layer and a $1\times 1$ convolutional layer to directly detect targets,
i.e., output binary maps whose size is equal to the size of original infrared images, with ``1'' representing the pixel belonging to targets and ``0'' representing the target belonging to backgrounds.
The results can be shown in the first line of Tab. \ref{tab:Ablation-study-on} (DBR\_v0).
Using DSW alone to detect targets, F1-score is only 23.88\%, which is lower compared with our DBR (64.10\%).
This is because the final purpose of ISTD is to determine whether each pixel belongs to the target, in other words, ISTD is a pixel-to-pixel task.
Due to the resolution of the feature map of DSW is gradually decrease, DSW is not suitable to pixel-to-pixel tasks.

\subsection{MAE Promotes Performance under Complex Backgrounds}

In traditional ISTD methods, background reconstruction, background modeling, or background estimation belong to the same kind of technology, which has long been studied by scholars and belongs to a relatively classical technology.
From the information theory, the ideal feature extractor decomposes the image information $H ({\rm Image})$ consists of two parts \cite{informationthery}:
\begin{equation}
H ({\rm Image}) =H({\rm Innovation})+H ({\rm Prior\ Knowledge}),
\end{equation}
where $H({\rm Innovation})$ denotes the novelty part, say small targets to be detected; and $H ({\rm Prior\ Knowledge})$ is the redundant information, say backgrounds which should be suppressed by background-reconstruction-based methods. 

The background-reconstruction-based methods had been proven effective for small target detection when the detector is stationary or the background is simple.
On the contrary, if the background is complex and the clutter interference is serious, traditional background-reconstruction-based ISTD methods would fail.
That is because when the background is complex, the information of the background does not only in $H ({\rm Prior\ Knowledge})$, but also in $H({\rm Innovation})$, resulting in the target in $H({\rm Innovation})$ cannot be distinguished from the background.

\begin{figure*}[!t]
\begin{centering}
\includegraphics[width=0.8\textwidth]{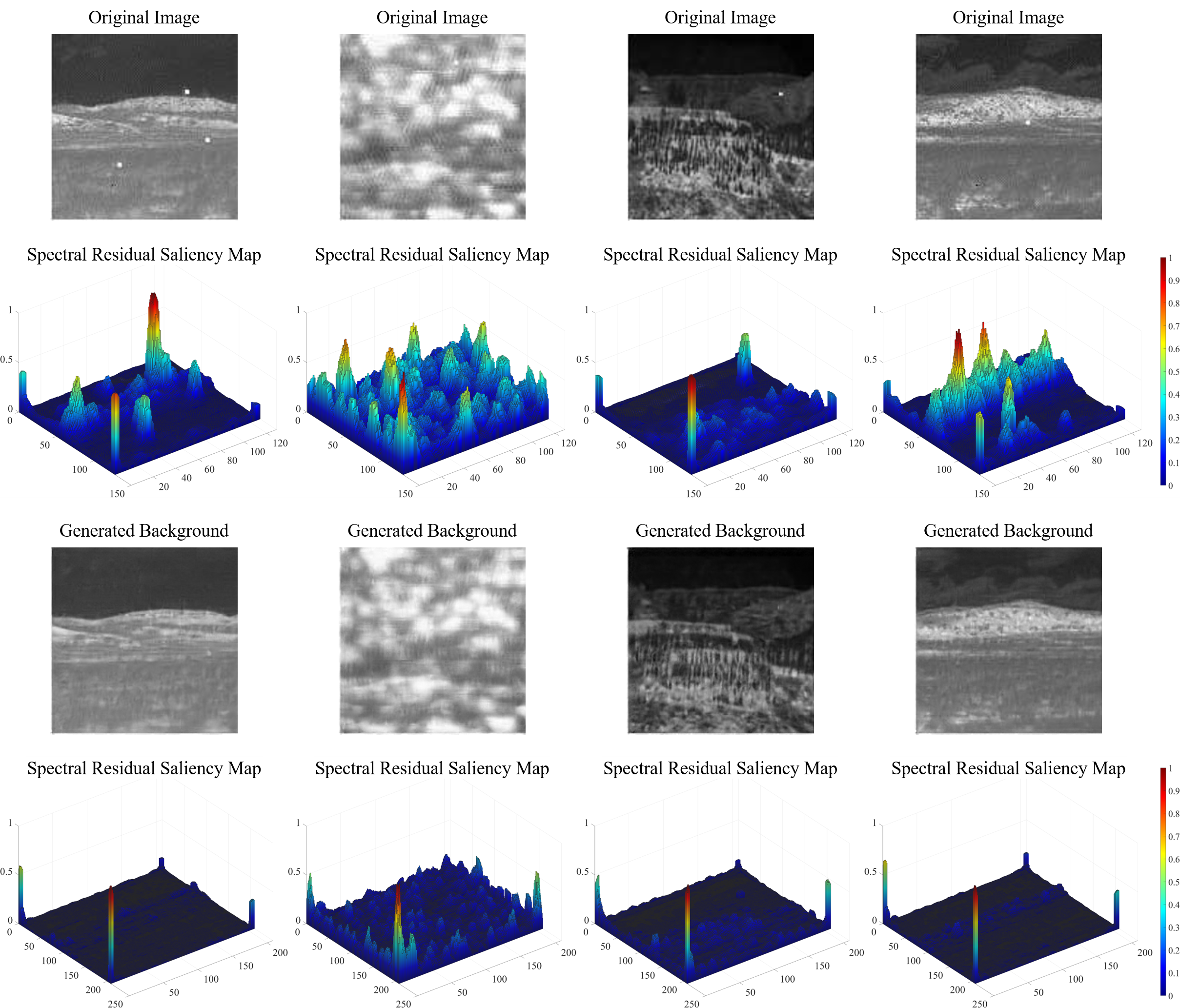}
\par\end{centering}
\caption{\label{fig:10}The spectral residual saliency maps of original infrared images and generated backgrounds.}
\end{figure*}

Deep-learning-based methods, however, have strong representation abilities, which can encode background information into $H ({\rm Prior\ Knowledge})$ as much as possible.
And $H({\rm Innovation})$ only contains target information, so the targets can be distinguished from backgrounds.
To this end, we select several infrared images with complex backgrounds and plot the spectral residual (SR) saliency maps of original images and generated backgrounds, as shown in Fig. \ref{fig:10}.
From the figure, we can find that in the original images, not only targets but also parts of backgrounds are prominent.
Directly using original images cannot distinguish targets from backgrounds.
But in the backgrounds generated by MAE, SR saliency maps are quite smooth and small compared with that of original infrared images, which is beneficial to the ISTD problem.

\subsection{Ablation Study}

\begin{table}[!h]
\centering{}\caption{\label{tab:Ablation-study-on}Ablation study on the MFIRST dataset.
``$\checked$'' indicates using the corresponding module, while ``$\times$'' indicates not.}
\resizebox{0.5\textwidth}{!}{
\begin{tabular}{cccccccccc}
\hline 
\multirow{2}{*}{Methods} & \multirow{2}{*}{DSW} & \multirow{2}{*}{BR} & \multirow{2}{*}{DH} & \multirow{2}{*}{Precision} & \multirow{2}{*}{Recall} & \multirow{2}{*}{F1} & Flops & Params & \multirow{2}{*}{FPS}\tabularnewline
 &  &  &  &  &  &  & (GMac) & (M) &  \tabularnewline
\hline 
DBR\_v0 & $\checked$ & $\times$ & $\times$ & 15.91 & 80.41 & 23.88 & 3.26 & 21.62 & 76.94\tabularnewline
DBR\_v1 & $\times$ & $\checked$ & $\times$ & 27.72 & 32.53 & 20.91 & 70.30 & 329.24 & 26.39\tabularnewline
DBR\_v2 & $\checked$ & $\checked$ & $\times$ & 36.34 & 34.78 & 25.34 & 73.56 & 350.86 & 23.42\tabularnewline
DBR\_v3 & $\times$ & $\times$ & $\checked$ & 57.37 & 70.71 & 58.66 & 5.61 & 34.26 & 55.56\tabularnewline
DBR\_v4 & $\times$ & $\checked$ & $\checked$ & 62.04 & 71.10 & 62.63 & 75.91 & 363.50 & 21.69\tabularnewline
\hline 
DBR & $\checked$ & $\checked$ & $\checked$ & 63.79 & 72.24 & 64.10 & 79.17 & 385.12 & 19.96\tabularnewline
\hline 
\end{tabular}}
\end{table}

\begin{figure*}[!t]
\begin{centering}
\includegraphics[width=1\textwidth]{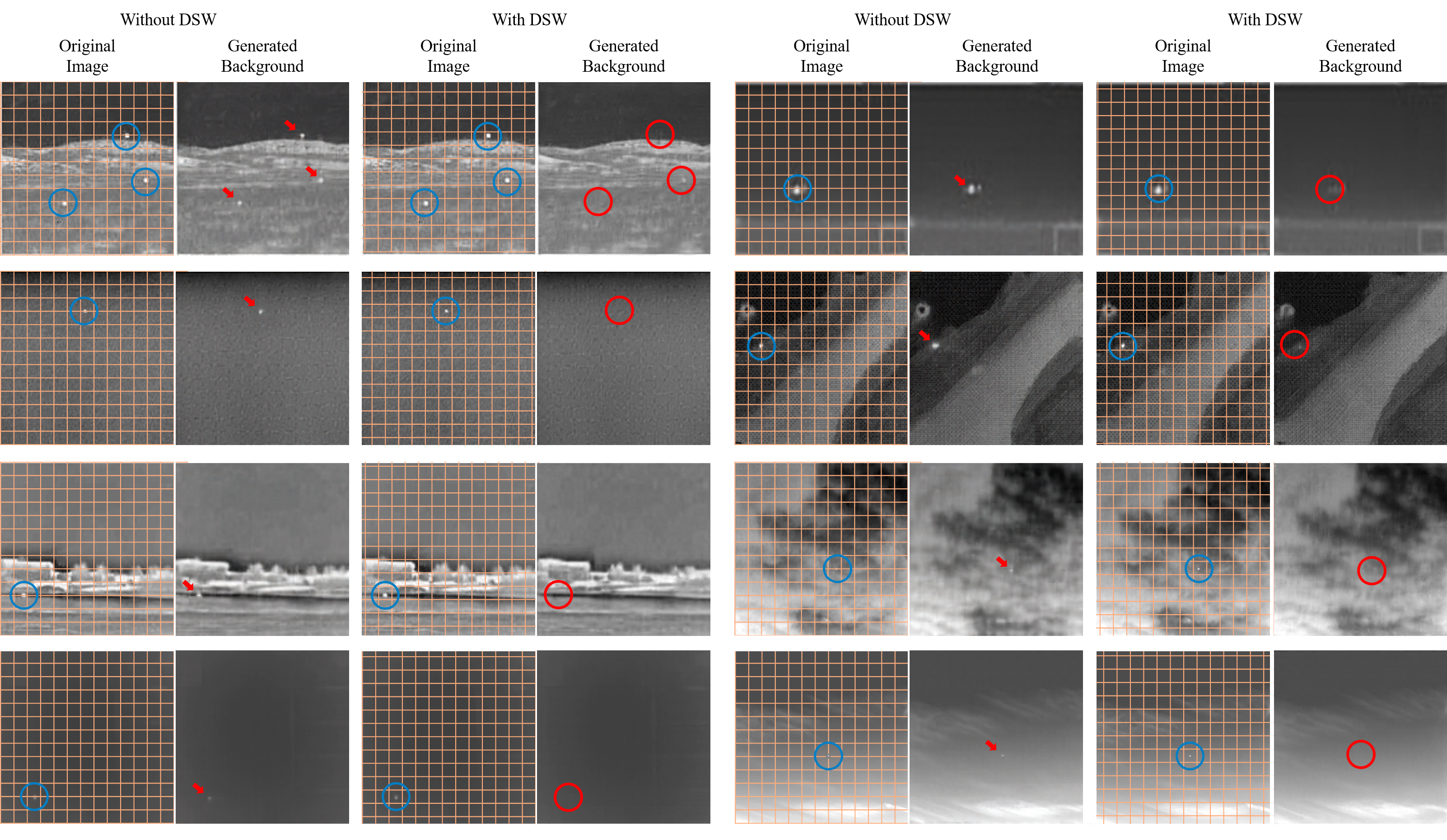}
\par\end{centering}
\caption{\label{fig:7}The performance comparison of background generation without DSW and with DSW.}
\end{figure*}

In order to demonstrate the effectiveness of our major modules, we implement three variations, separately pruning DSW, BR, or DH.
The comparison results on MFIRST dataset are shown in Tab. \ref{tab:Ablation-study-on}.
Among them, DSW cannot exist with DH, so we excluded this method in Tab. \ref{tab:Ablation-study-on}.

Compared DBR\_v1 with DBR\_v2, DSW improves F1-score by 4.43\%, compared DBR\_v4 with DBR, DSW improves F1-score by 1.47\%.
This is because, without DSW, BR is prone to divide one target into different patches, resulting in background reconstruction failure, which affects the detection performance.
To visually demonstrate the impact of DSW on background reconstruction performance,  we separately generate backgrounds with DSW and without DSW, as shown in Fig. \ref{fig:7}.
When the target is divided into different patches, directly reconstructing backgrounds will lead to targets also being reconstructed (the second column).
With DSW, the target can be divided into one patch. Due to targets do not belong to backgrounds, targets cannot be reconstructed (the last column).

Compared DBR\_v3 with DBR\_v4, BR improves F1-score by 3.97\%, which means that BR can improve detection performance under complex backgrounds.
Small and dim infrared targets are difficult to distinguish from backgrounds under complex backgrounds.
With BR, targets can be easily distinguished by the difference between original images and generated backgrounds.

Compared DBR\_v2 with DBR, DH improves F1-score by 38.76\%.
This is because the generated background is not absolutely the same as the factual background.
If the differences between original images and generated backgrounds are directly used as results, the detector will regard reconstruction errors as targets, reducing the detection performance.
Our proposed DBR achieves the best F1-score of 64.10\%, which means that the combination of DSW, BR, and DH can achieve the best detection performance.

As for computational cost, DSW and DH have 4.12\% and 7.09\% of all Flops; 5.61\% and 8.90\% of all Params, respectively.
Thus most of the computational cost comes from BR.
This is because Considering the reconstruction quality will seriously affect the detection performance; simple and small reconstruction models will lead to significant reconstruction errors.
Therefore, we utilize a mature but large Vision Transformer (MAE-ViT-large \cite{MAE}) for background reconstruction.
Given the best performance of DBR and the detection speed (19.96 FPS), this computational cost is acceptable.

\subsection{Comparison with Other Methods}

We compare DBR with traditional methods (WSLCM \cite{WSLCM}, TLLCM \cite{TLLCM}, ADMD \cite{ADMD}, MSPCM \cite{MSPCM}, and AADCDD \cite{ADDCDD}) and DL-based methods (MDvsFA-cGAN \cite{MDvsFAcGAN}, ALCNet \cite{ALCNet}, ACM \cite{ACM}, DNANet \cite{DNANet}, CourtNet \cite{CourtNet}, and IAANet \cite{IAANet}).
Wang et al.\cite{MDvsFAcGAN} emphasize that F-measure indicates good performance rather than only achieving high precision or recall rates.
DBR gets the best results for F1-score in two datasets over other methods, as shown in Tab. \ref{tab:Comparison-of-different}.
In specific, DBR achieves 64.10\% and 75.01\% for F1-score, which is 0.2\% and 1.75\% higher than the second-best method (IAANet).
Among DL-based methods, DBR has the closest precision rate and recall rate, indicating the effectiveness of the proposed weighted dice loss in balancing the precision rate and the recall rate.

\begin{table*}[!t]
\centering{}\caption{\label{tab:Comparison-of-different}Comparison of different methods which were evaluated on MFIRST and SIRST.}
\resizebox{\textwidth}{!}{%
\begin{tabular*}{1\textwidth}{@{\extracolsep{\fill}}ccccccc}
\hline 
\multirow{2}{*}{Methods} & \multicolumn{3}{c}{MFIRST} & \multicolumn{3}{c}{SIRST}\tabularnewline
\cline{2-7} \cline{3-7} \cline{4-7} \cline{5-7} \cline{6-7} \cline{7-7} 
 & Precision & Recall & F1 & Precision & Recall & F1\tabularnewline
\hline 
ADMD \cite{ADMD} & 40.75 & 44.21 & 35.77 & 57.09 & 69.03 & 54.35\tabularnewline
MSPCM \cite{MSPCM} & 49.23 & 49.36 & 39.07 & 69.17 & 69.23 & 62.80\tabularnewline
AADCDD \cite{ADDCDD} & 50.16 & 56.19 & 42.12 & 78.94 & 62.63 & 64.04\tabularnewline
TLLCM \cite{TLLCM} & 57.70 & 46.11 & 45.15 & 78.19 & 59.08 & 61.18\tabularnewline
WSLCM \cite{WSLCM} & 68.66 & 61.40 & 58.08 & 74.86 & 71.89 & 66.72\tabularnewline
\hline 
MDvsFA-cGAN \cite{MDvsFAcGAN} & 66.00 & 54.00 & 60.00 & - & - & -\tabularnewline
ALCNet \cite{ALCNet} & - & - & - & 77.98 & 69.02 & 69.97\tabularnewline
ACM \cite{ACM} & - & - & - & 67.09 & 85.02 & 66.76\tabularnewline
DNANet\cite{DNANet} & 56.82 & 70.83 & 57.89 & 61.43 & 90.67 & 70.99\tabularnewline
CourtNet\cite{CourtNet} & 60.87 & 72.61 & 61.80 & 69.60 & 83.33 & 72.81\tabularnewline
IAANet\cite{IAANet} & 60.60 & 81.78 & 63.90 & 69.25 & 87.98 & 73.26\tabularnewline
DBR (Ours) & 63.80 & 72.24 & \textbf{64.10} & 80.70 & 74.09 & \textbf{75.01}\tabularnewline
\hline 
\end{tabular*}}
\end{table*}

\subsection{Qualitative Performances}

\begin{figure*}[!t]
\begin{centering}
\includegraphics[width=1\textwidth]{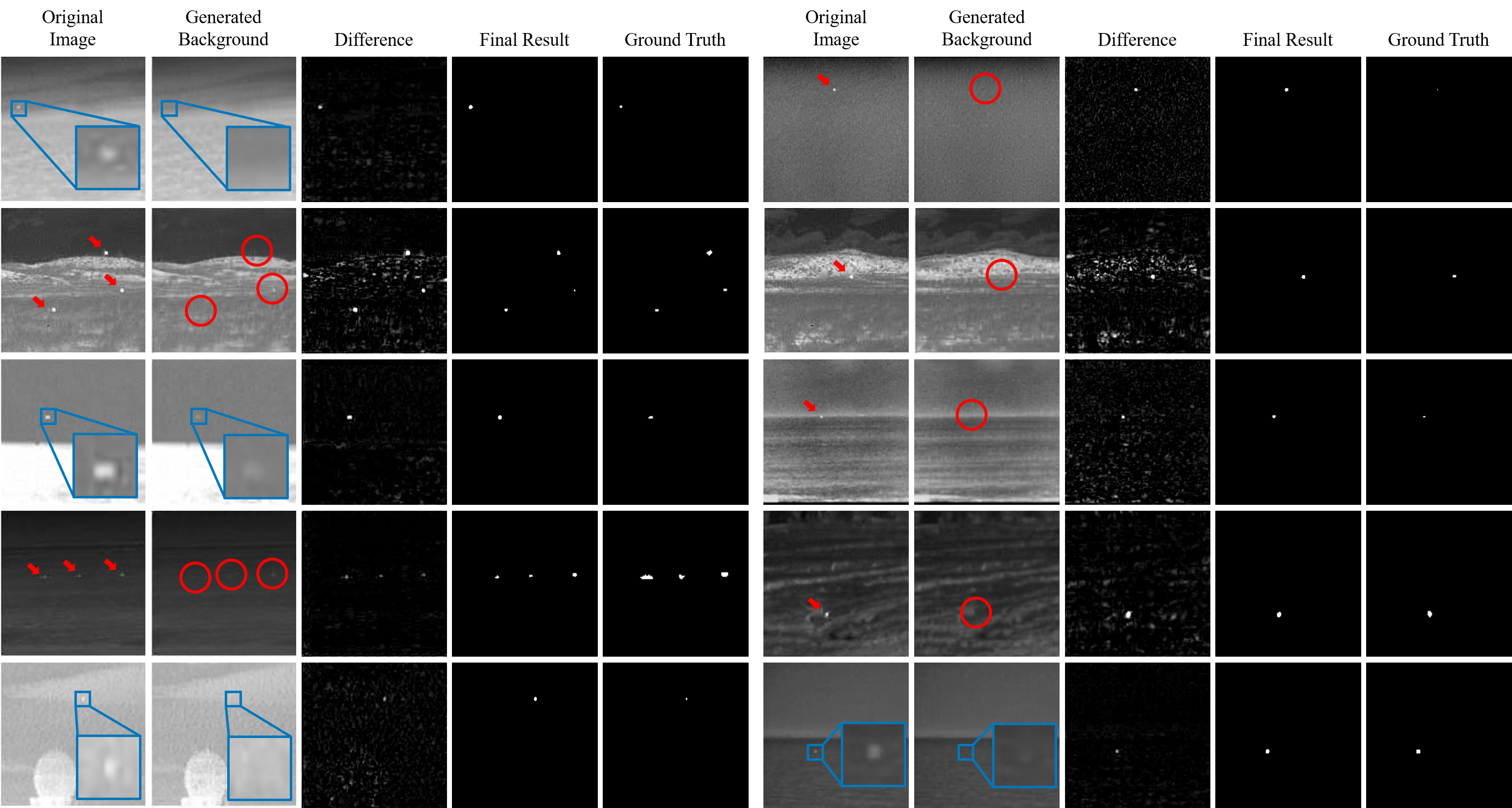}
\par\end{centering}
\caption{\label{fig:6}Qualitative performances of DBR.}
\end{figure*}

In order to illustrate the detection performance of DBR, as well as whether BR could generate clean backgrounds, we select 10 samples from MFIRST for comparison.
Fig. \ref{fig:6} shows infrared images (the first column), generated backgrounds (the second column), their differences (the third column), final results (the fourth column), and ground truth (the last column). 

In the first column, targets in the original infrared images are dim and small, which is difficult to distinguish from the complex backgrounds.
In the second column, BR can generate clean backgrounds without targets.
In the differences between the original images and generated backgrounds (the third column), the target is more apparent than in the original images, which means that background reconstruction benefits the ISTD task.
However, there are lots of noises in images of the third column caused by reconstruction errors, so the third column cannot be directly used as the detection result.
DH filters these noises, which demonstrates the effectiveness of reducing the impact of reconstruction errors on detection performance.

Based on all the experiments performed in this section, we conclude that:
\begin{enumerate}
\item Compared with the existing ISTD methods (WSLCM \cite{WSLCM}, TLLCM
\cite{TLLCM}, ADMD \cite{ADMD}, MSPCM \cite{MSPCM}, and AADCDD
\cite{ADDCDD}, MDvsFA-cGAN \cite{MDvsFAcGAN},
ALCNet \cite{ALCNet}, ACM \cite{ACM}, DNANet \cite{DNANet}, CourtNet
\cite{CourtNet}, and IAANet \cite{IAANet}), DBR achieves the best F1-score on the two ISTD datasets, MFIRST (64.10\%) and SIRST (75.01\%), and its speed is 19.96 FPS.
\item BR can generate clean backgrounds without targets. The performance of the detector with BR is better than that of the detector without BR. Background reconstruction benefits ISTD.
\item With the consideration of dynamically shifting windows, DSW can avoid dividing one target into two patches. DSW benefits background reconstruction.
\item DH can filter noises in the differences between original images and generated backgrounds. DH makes the detector robust to reconstruction errors.
\end{enumerate}

\section{Conclusion}

In this paper, we detect infrared small targets under complex backgrounds by background reconstruction.
We propose a novel ISTD method called Dynamic Background Reconstruction (DBR). 
The principle of DBR is that targets can be detected through infrared images with targets minus clean backgrounds without targets. 
To this end, we propose a background reconstruction module (BR) based on Transformer (MAE).
BR reconstructs twice grid-masked images with a masking ratio of 50\%. 

If one target was divided into two neighboring patches, BR would fail to reconstruct a clean background. 
To solve this problem, we propose a dynamic shift window module (DSW). 
DSW calculates offsets with that infrared images dynamically shift before input embedding. 
So that can prevent DBR divide one target into different patches. 

In order to reduce the influence of reconstruction error on detection performance, we propose a detection head (DH) and a weighted dice loss (WDLoss).
DH utilizes a structure of the densely connected Transformer.
Original images, generated backgrounds, and their differences are fed into DH to get the final results.
In the training phase, WDLoss balances the precision and recall rates.
Extensive experiments and ablation studies demonstrate the effectiveness of the proposed main modules. 

In the future, we will design a dedicated background reconstructor for ISTD.
We will design the reconstructor as a ``plug-and-play'' module and try integrating BR into other ISTD methods to improve the detection performance.

{\small
\bibliographystyle{ieee_fullname}
\bibliography{ref}
}

\end{document}